\documentclass[letterpaper, 10 pt, conference]{ieeeconf}  

\IEEEoverridecommandlockouts                              

\usepackage{graphicx}
\usepackage{amsmath}
\usepackage{amssymb}

\title{\LARGE \bf Bayesian deep learning for affordance segmentation in images}

\author{Lorenzo Mur-Labadia, Ruben Martinez-Cantin and Jose J. Guerrero
\thanks{The authors belong to the Instituto de Ingenieria de Aragon (I3A), Universidad de Zaragoza, España {\tt\small lmur@unizar.es}, {\tt\small rmcantin@unizar.es}, {\tt\small josechu.guerrero@unizar.es}. This work was supported by the projects PID2021-125209OB-I00 and TED2021-129410B-I00 (MCIN/AEI/10.13039/501100011033 and NextGenerationEU/PRTR)}}

\begin{document}

\maketitle
\thispagestyle{empty}
\pagestyle{empty}

\begin{abstract}

Affordances are a fundamental concept in robotics since they relate available actions for an agent depending on its sensory-motor capabilities and the environment. We present a novel Bayesian deep network to detect affordances in images, at the same time that we quantify the distribution of the aleatoric and epistemic variance at the spatial level. We adapt the Mask-RCNN architecture to learn a probabilistic representation using Monte Carlo dropout. Our results outperform the state-of-the-art of deterministic networks. We attribute this improvement to a better probabilistic feature space representation on the encoder and the Bayesian variability induced at the mask generation, which adapts better to the object contours. We also introduce the new Probability-based Mask Quality measure that reveals the semantic and spatial differences on a probabilistic instance segmentation model. We modify the existing Probabilistic Detection Quality metric by comparing the binary masks rather than the predicted bounding boxes, achieving a finer-grained evaluation of the probabilistic segmentation. We find aleatoric variance in the contours of the objects due to the camera noise, while epistemic variance appears in visual challenging pixels. 

\end{abstract}

\section{Introduction}

Defined by the psychologist J.J Gibson \cite{gibson}, affordances are the different action possibilities available in the environment. They relate the objects, the possible actions, the context and the posterior consequences of the selected action. From a robotics perspective, the concept of affordances \cite{jamone2016affordances} emerges as a powerful tool for encoding the object properties for two reasons. First, affordances depend on perceptual, and motor capabilities and the learning strength of the agent. Second, the action possibilities encoded in the affordance definition are the basis for the posterior prediction of the action consequences and future path planning. For example, a door is \textit{openable} or \textit{closeable}, but it depends on its current state (the context) and the ability of the agent to grasp the doorknob (motor capabilities), while it determines the future state of the object (action possibilities). 

Deep learning models for computer vision allow robots to capture a lot of information from the scene. However, to treat these methods like other sensors, it is required that they estimate the uncertainty. It allows, using the established Bayesian techniques, to fuse the network’s predictions with prior knowledge or other sensor measurements, or to accumulate information over time. Current instance and affordance segmentation models produce \textit{deterministic} predictions, which are not calibrated probabilities, and which cannot be adapted to a Bayesian sensor fusion framework. We extend a popular instance segmentation model to produce \textit{probabilistic} detections by the application of a Bayesian deep learning technique, Monte-Carlo dropout (MC-dropout). Furthermore, we compute the contribution of the aleatoric and epistemic variance, showing that the estimation of the spatial uncertainty allows a deeper affordance reasoning and extracts more information. The uncertainty estimation is very relevant in safety-critical applications. For example, as Figure \ref{fig:segm} shows, we appreciate that the most uncertain part of the knife handle is the one closer to the edge, which is also the most dangerous to manipulate. Our main contributions are:

\begin{figure}
\centering
\includegraphics[width=0.49\columnwidth]{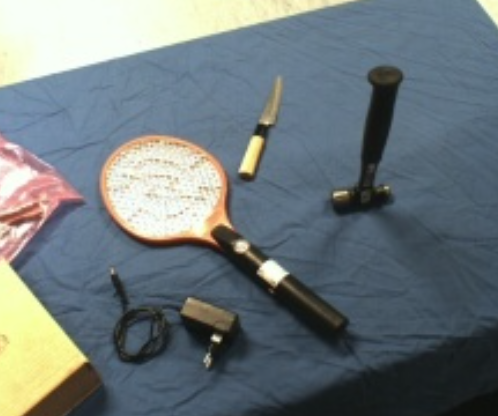}
\includegraphics[width=0.49\columnwidth]{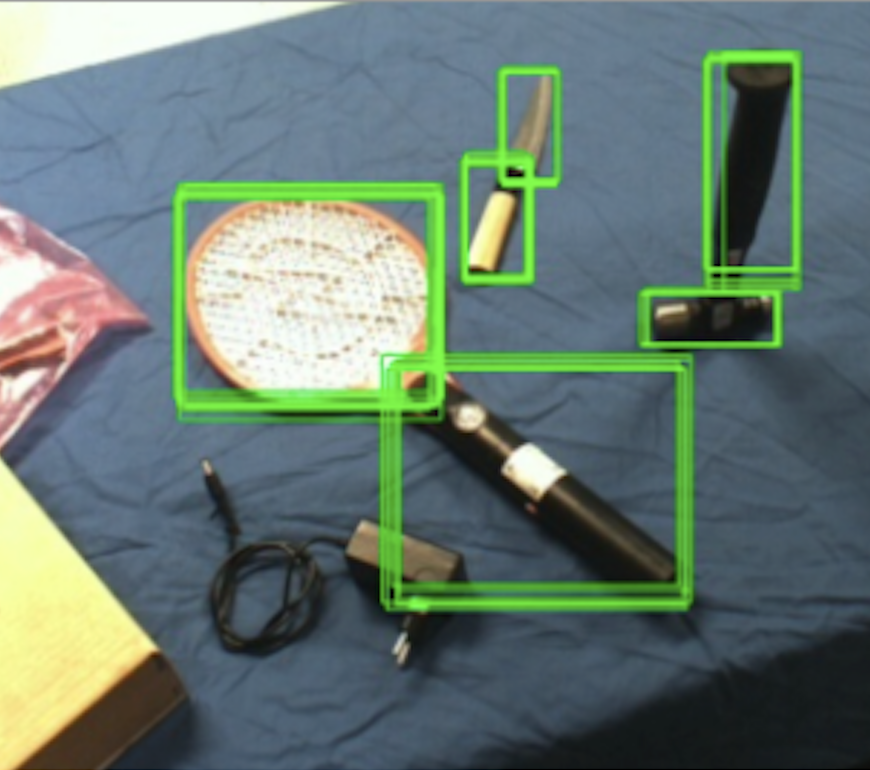} \\
\includegraphics[width=0.49\columnwidth]{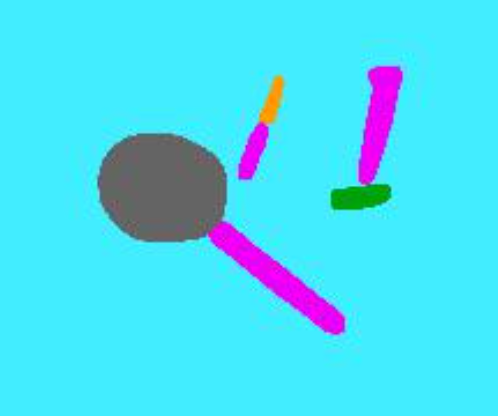}  
\includegraphics[width=0.49\columnwidth]{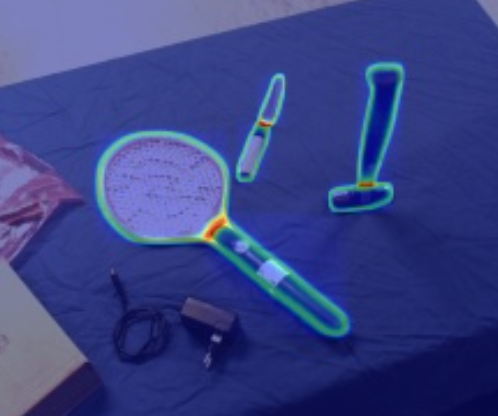}  
\caption{We detect the affordances in an image (bottom-left) from a RGB image (top-left). Our Bayesian Instance Segmentation architecture produces a probabilistic distribution of the mask and class label, computing the respective epistemic and aleatoric variance. Despite previous approaches that were limited to a probabilistic detection (top-right), our Our Bayesian Instance Segmentation architecture produces per-pixel variance map (bottom-right).}
\label{fig:segm}
\end{figure}

\begin{itemize}
    \item We extend affordance segmentation to a probabilistic stage, extracting a per-pixel estimation of the aleatoric and epistemic variance at the spatial level.
    \item We analyse the effect of MC-dropout on the architecture and report several uncertainty quality metrics.
    \item We achieve state-of-the-art performance on the IIT-Aff dataset, showing that properly calibrated Bayesian models outperform their respective deterministic version.
    \item We introduce the Probability-based Mask Quality measure (PMQ), an extension of the  Probability-based Detection Quality measure (PDQ), which evaluates the probabilistic predicted masks.
\end{itemize}

\section{Related works}

Perception in robotics was boosted by deep learning models since they can capture a lot of information from the scene. On the affordance detection task, multiple approaches with different architecture proposals emerged: Nguyen et al. \cite{nguyen2016detecting} applied an encoder-decoder to learn the affordances from the deep latent space, \cite{nguyen2017object} post-processed the feature maps with Conditional Random Fields to refine the class boundaries, Affordance-Net \cite{do2018affordancenet} modified the Faster R-CNN object detector with more deconvolutional layers to upsample the RoI feature maps, and the authors in \cite{minh2020learning} increased the performance with the use of more powerful backbones and the incorporation of multiple alignments. Finally, \cite{caselles2021standard} proposed to reuse a standard object detection model for the task of affordance perception. Their results show that predicting the affordance without an intermediate object detection is more efficient than decoupling into predicting first the object and then extracting their different affordances.

On the other side, Bayesian deep learning allows the quantification of uncertainty by large neural networks. Since the analytical computation of the posterior distribution is intractable, several approximations have attempted this problem. MC-dropout approximates the posterior as the sample distribution of the $M$ forward passes with the dropout layers active during the inference stage \cite{Gal}. On the other hand, deep ensembles \cite{gustafsson2020evaluating} consists in training an ensemble of $M$ models with random initialization of their neural network parameters and shuffling of the dataset. Kendall and Gal \cite{Gal} introduce uncertainty estimation with a Bayesian neural network in semantic segmentation and depth regression tasks. They show that while the aleatoric uncertainty appears in the contours of objects and far away regions, epistemic uncertainty appears in challenging pixels where the input sample is out-of-distribution of the training data. For object detection, sampling-based approaches average the output of $M$ different detection models to extract an estimation of the uncertainty distribution with multiple uses: the spatial uncertainty estimation in the bounding boxes of the detected vehicles reflects the complex environmental noises in LiDAR perception \cite{feng2021labels}, epistemic variance rejects false positive detections of object classes not present in the training dataset \cite{miller2021uncertainty} and Bayes-OD \cite{harakeh2020bayesod} avoids the loss of information by substituting the non-Maximum suppression components with a Bayesian treatment. While most works were limited to object detection, sampling-based instance segmentation \cite{morrison2019uncertainty} proposes to group the multiple mask detections to produce a fine-grained mask around the detected object. However, its estimation of the spatial uncertainty reduces it to a single value and losses all the pixel-wise distribution.

\section{Bayesian affordance segmentation}

Based on the previous work of Caselles et al. \cite{caselles2021standard}, we adopt an instance segmentation model (Mask-RCNN) for our affordance segmentation task. Mask-RCNN \cite{he2017mask}, developed on top of Faster R-CNN, is a region-based CNNs that returns the class label, bounding boxes and a binary mask with a confidence score for each instance. The first stage of Faster R-CNN is an encoder followed by a Region Proposal Network that gives a set of Region Proposals (RP) based on the encoder features. It predicts potential bounding boxes $b = \{x_1, y_1, x_2, y_2\}$ with an object inside and then classifies the objects in each bounding box separately with a softmax layer to output a vector $p_c$ of $c+1$ class probabilities. A RoIAlign layer fixes the size of the proposed Regions of Interest (RoI) and refines the bounding boxes and the object class of each RP. Then, the mask head applies a few convolution layers to generate a binary mask $\Upsilon$ for each RoI and segments the image at the pixel level. We also keep the sigmoid mask vector, which we normalize to obtain the probability mask vector $p_\Upsilon(x)$. At inference, each forward pass of Mask-RCNN produces a set of individual detections $S = \{ D_1, ..., D_k \}$ where each detection is $D_i = \{b, p_c, \Upsilon, p_\Upsilon(x) \}$. 

Our probabilistic instance segmentation is divided into three stages. First, we adapt Mask-RCNN incorporating intermediate dropout layers to produce a large number of $D_i$ detections using MC-dropout. Then, we adopt a merging strategy to group $D_i$ detections in $\mathcal{O}_i$ observations, where the $N$ merged detections provide the probability distribution of a single observation $\mathcal{O}_i = \{ D_1,...,D_N\}$. Finally, from the probability distribution, we extract the uncertainty estimation and deterministic metrics to evaluate the quality of the predictions.

\begin{figure}
\centering
\includegraphics[width=0.99\columnwidth]{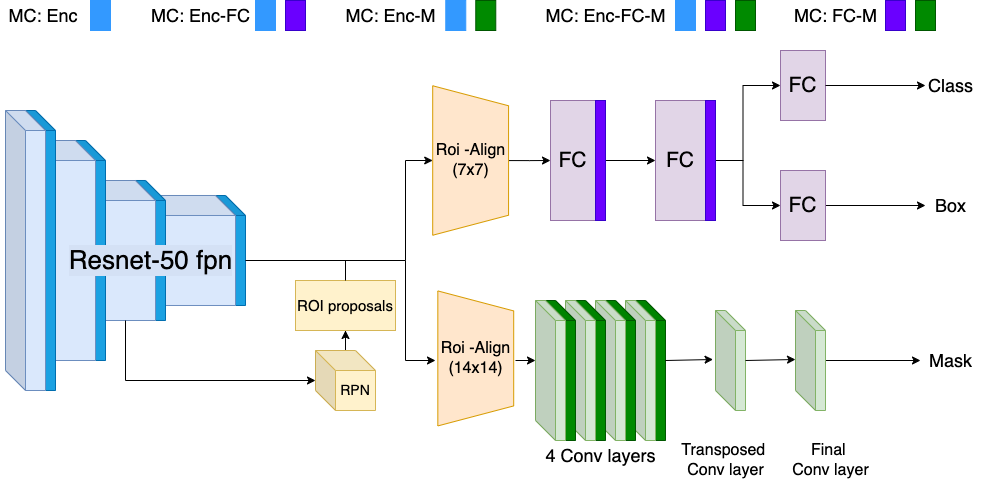}
\caption{Different configurations of MC-dropout in Mask-RCNN. We place dropout layers at the end of the four-layer blocks of the backbone (light-blue), after the activation layer of the four convolutional layers in the mask branch (green) and after the activation layer of the two fully-connected layers of 1024 channels in the box branch (purple).}
\label{fig:maskrcnn}
\end{figure}

\subsection{MC-Dropout}

Sampling-based techniques produce $M$ samples $\mathbb{S} = \{S_1, ..., S_m\}$, where each sample $S_j = \{D_1, ..., D_k \}$ is composed by a set of detections $D_i$. Using MC-dropout, we approximate the true posterior of the observation as the sample distribution of the $N$ different detections that form the observation. It is a special case of variational inference, where the Kullback-Leibler divergence of the approximate posterior to the true posterior is minimized by approximating the variational parameters of a Bernoulli distribution \cite{gal2016dropout}. Inspired by Rodriguez et al. \cite{rodriguez2022bayesian}, we create different configurations and compare the effects of the different dispositions of the dropout layers to measure the effect of the Bayesian component in each part, as shown in Figure \ref{fig:maskrcnn}. 

\subsection{Merging object masks}

Once we compute all the detections $D_i$, we group them to form the observations $\mathcal{O}_i$ of affordances in the scene. Following Miller et al. \cite{miller2019evaluating}, we define observation as a group of detections with a high (spatial and semantic) affinity between them. They compare the performance of different affinity metrics and merging strategies, showing that the Basic Sequential Algorithm Scheme (BSAS) and the Intersection over Union (IoU) of the combination of the detected masks $\Upsilon$ achieve the best performance. We only grouped samples with the same label class, which avoids the merging of overlapped objects whose masks were very close and ensured that the observation belongs to the same class. Compared with previous approaches limited to object detection \cite{miller2018dropout, miller2019evaluating}, the advantage of comparing pixel-wise masks in our probabilistic instance segmentation rather than bounding boxes is that two significantly overlapping masks are much less likely to represent the same object than two overlapping bounding boxes, especially in the case of many tightly grouped or irregularly shaped objects. 

\subsection{Spatial uncertainty estimation}  

To obtain a discrete output from the probability distribution, we average the \textit{multiple detected} bounding boxes coordinates $b$, the class probability $p_c$ and the masks probabilities $p_\Upsilon(x)$ to form a \textit{single observed} bounding box $\mathcal{O}(b)$, class probability $\mathcal{O}(p_c)$ and mask $\mathcal{O}(p_\Upsilon(x))$. The variance of the predictive distribution is the sum of the epistemic and aleatoric uncertainty. Epistemic uncertainty  $\sigma_e(x)$, related to the model knowledge, is reduced as the dataset increases and covers more variability. It represents a probability distribution over the model parameters. Aleatoric uncertainty $\sigma_a(x)$ reflects the noise in the observations like the camera motion or object boundaries. It is not reduced by collecting more data as it is inherent to the data distribution. It encodes the variability in the respective inputs from the test data. Previous works \cite{morrison2019uncertainty, miller2018dropout, miller2019evaluating} ignore the differences between these two uncertainties and reduce the spatial uncertainty to a single scalar value. Following \cite{Gal, kwon2020uncertainty}, we present a unified approach that computes the spatial uncertainty $\sigma_{sp}(x)$ as the sum of the epistemic and aleatoric variance. We extend the probabilistic object detection to the novel probabilistic instance segmentation since it extracts the fine-grained uncertainty at the pixel level from the detected mask outputs $p_\Upsilon(x)$ using Equation 1.

\begin{equation}
   \sigma_{sp}(x) = \underbrace{\frac{1}{N} \sum_{n=1}^{N} \textrm{diag}({p_n}) - {p_n^{\otimes^2}}}_{aleatoric} + \underbrace{\frac{1}{N} \sum_{n=1}^{N} ({p_n} - \bar{p})^{\otimes^2}}_{epistemic} 
\end{equation}

\noindent where $p_n$ is the mask probability vector of the $n$ detection $p_n = p_\Upsilon(x)_n$, $\bar{p}$ is the mean of the probability vectors $\bar{p} = \frac{1}{N} \sum_{n = 1}^N p_n$, the $p^{\otimes^2}$ operator multiplies the vector itself by its transpose $p^{\otimes^2} = p \cdot p^T$ and $N$ is the number of detections that form the observation. Note that, as a future work, we can extend this equation to obtain the semantic uncertainty by substituting $p_n$ by the class vector $p_c$ of the $n$ detection.

\section{Experimental setup}

We train our model with a Stochastic Gradient Descent (SGD) optimizer and learning rates of $10^{-2}$ and $10^{-5}$ for Resnet-50 and Resnext-101, respectively. We decrease the learning rate to avoid the issue of exploding gradient. Following previous procedures \cite{caselles2021standard}, we modify the number of predictions heads to the number of affordance classes plus the background class $c + 1$, and we use pre-trained versions on the COCO datasets \cite{lin2014microsoft} of the backbones. In both cases, we start the training with a linear warm-up during the first 500 epochs from $10^{-3}$ and $10^{-6}$, respectively. For the BSAS clustering algorithm with IoU and Same Label conditions, we set as hyper-parameters two minimum detections to form an observation, a 0.5 softmax score and a 0.5 IoU spatial affinity threshold, following other works \cite{miller2019evaluating}. 

\subsection{Datasets}

We perform our experiments on the IIT-AFF dataset \cite{nguyen2017object}, a robotics-based dataset that contains 8,835 images in real-world scenarios. Compared to the controlled scenario or synthetic datasets \cite{thermos2017deep, myers2015affordance}, the IIT-Aff contains annotations in real-world environments with contextual information and large variability. It covers 7 different affordances categories (\textit{contain, cut, display, engine, grasp, hit, pound, support, w-grasp}), which represent common manipulation capabilities of 10 different objects (\textit{bottle, bowl, cup, drill, hammer, knife, monitor, pan, racket, spatula}).

\subsection{Metrics}

Following previous procedures \cite{caselles2021standard, minh2020learning, do2018affordancenet}, we use the $F_\beta^w$ score \cite{margolin2014evaluate} to compare the performance of the deterministic predictions with previous baselines and show our improvement with the state-of-the-art. This metric evaluates the foreground maps and weights the error of the pixels to correct three assumptions: dependency, interpolation and equal importance. In the equation, $P_w$ and $R_w$ corresponds to the weighted Precision and Recall and the hyper-parameters $\beta = 1$, $\sigma^2 = 4$, $\alpha = \frac{(0.5)}{5}$ are the same as in previous works.

\begin{equation}
    F_\beta^w = (1 + \beta^2) \cdot \frac{P_w \times R_w}{\beta^2 \cdot P_w + R_w}
\end{equation}

\begin{table*}[t]
\centering
\resizebox{\textwidth}{!}{%
\begin{tabular}{c|cccccccc}
\hline
 & \begin{tabular}[c]{@{}c@{}}ED\\  RGB \cite{nguyen2016detecting} \end{tabular} & \begin{tabular}[c]{@{}c@{}}BB \\ CNN \cite{nguyen2017object} \end{tabular} & \begin{tabular}[c]{@{}c@{}}AN\\  VGG 16 \cite{do2018affordancenet} \end{tabular} & \begin{tabular}[c]{@{}c@{}}AN \\  SE154 \cite{minh2020learning} \end{tabular} & \begin{tabular}[c]{@{}c@{}} Mask \\ R-CNN R50 \cite{caselles2021standard} \end{tabular} & \begin{tabular}[c]{@{}c@{}}Bayesian \\ Mask-RCNN R50\end{tabular} & \begin{tabular}[c]{@{}c@{}}Mask R-CNN \\ Rx101\end{tabular} & \begin{tabular}[c]{@{}c@{}}Bayesian \\ Mask-RCNN Rx101\end{tabular} \\ \hline
contain & 66.4 & 75.6 & 79.6 & 81.9 & 83.6 & 86.4 & 85.8 & \textbf{86.5} \\
cut & 60.7 & 69.9 & 75.7 & 79.8 & 84.7 & 82.0 & \textbf{86.4} & 84.6 \\
display & 55.4 & 72.0 & 77.8 & 82.5 & 86.3 & 89.5 & 89.9 & \textbf{90.6} \\
engine & 56.3 & 72.8 & 77.5 & 83.2 & 88.9 & 89.4 & 90.5 & \textbf{90.9} \\
grasp & 59.0 & 63.7 & 68.5 & 77.1 & 71.1 & 77.1 & 77.5 & \textbf{78.2} \\
hit & 60.8 & 66.6 & 70.8 & 79.2 & 92.3 & 93.4 & 94.3 & \textbf{95.4} \\
pound & 54.3 & 64.1 & 69.6 & 78.2 & \textbf{81.8} & 80.6 & 80.5 & 80.7 \\
support & 55.4 & 65.0 & 69.8 & 78.9 & 86.7 & 85.5 & 85.4 & \textbf{88.7} \\
w-grasp & 50.7 & 67.3 & 71.0 & 80.6 & 83.7 & 85.8 & 84.7 & \textbf{86.6} \\ \hline
Average & 57.6 & 68.6 & 73.4 & 80.2 & 84.4 & 85.7 & 86.1 & \textbf{86.9} \\ \hline
\end{tabular}%
}
\caption{Affordance perception on the IIT-Aff Dataset ($F_\beta^w$ * 100 score ($\uparrow$)). We compare our Bayesian and deterministic models with previous works, achieving the best performance. We attribute the improvement due to the better generalization of Bayesian models and the higher modelling capacities of the Resnext-101 encoder in the latent space.}
\label{tab:baselines}
\end{table*}

The PDQ score \cite{hall2020probabilistic} evaluates the uncertainty estimation for probabilistic object detectors. Our novel Probability-based Mask Quality (PMQ) measure reports the quality of the probabilistic instance segmentation task. We adapt the PDQ to the new task by comparing the mask outputs instead of the bounding box predictions. The PMQ computes the label $Q_l$ and spatial $Q_s$ quality of a detector based on the class label vector $p_c$ and the mask $\Upsilon$, respectively. Compared with the mean Average Precision (mAP) score \cite{lin2014microsoft}, the PMQ does not rely on IoU thresholds that filter the number of detections and compute the optimal assignment between the detections and the ground truth using the spatial and label uncertainty quality. 

\begin{figure}[]
\centering
\includegraphics[width=0.99\columnwidth]{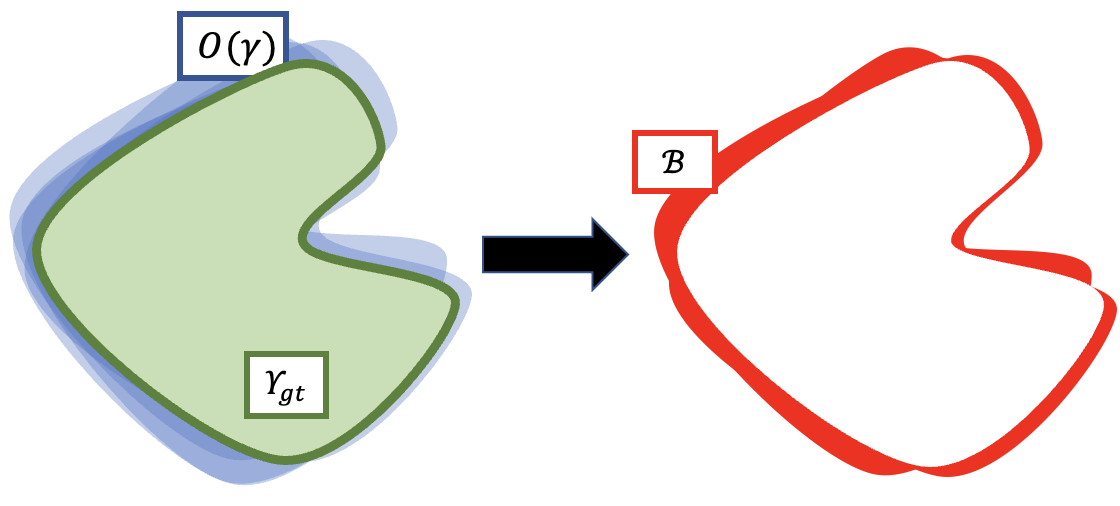}
\caption{Our extended PMQ re-defines the background evaluation region $\mathcal{B}$ as the difference between the grouped detected masks $O(\Upsilon)$ with the ground truth mask $\Upsilon_{gt}$.}
\label{fig:com_bay_det}
\end{figure}

\begin{equation}
Q_s(\Upsilon_{gt}, \Upsilon_{h}) = \exp(-(L_{FG}(\Upsilon_{gt}, \Upsilon_{h})) + (L_{BG}(\Upsilon_{gt}, \Upsilon_{h}))
\end{equation}

\noindent Our novel Spatial Quality $Q_s$ is the sum of the exponential of the foreground loss $L_{FG}$ and the background loss $L_{BG}$ at the mask level, as shows Equation 3. It scores 1 when all the ground truth mask $\Upsilon_{gt}$ pixels have assigned a spatial probability of 1 and the rest of the pixel's probabilities is 0. The foreground loss $L_{FG}$, computed as the negative log probability of the pixel's probability, penalizes those pixels inside the ground truth segment with low probability. Instead of computing a Gaussian heatmap using the bounding boxes as in the PDQ case, which loses the boundary detections, we use the masks $\mathcal{O}(\Upsilon)$ to obtain a fine-grained heat map $\Upsilon_{h}$. The value assigned to each pixel $x$ is the number of times that the pixel is detected $\Upsilon > 0.5$ divided by the total number of samples $M$. Therefore, as illustrates Figure \ref{fig:com_bay_det}, the background evaluation $\mathcal{B}$ is the difference between grouped detected masks $\mathcal{O}(\Upsilon)$ with the ground truth mask $\Upsilon_{gt}$.

Finally, we show two uncertainty error metrics, the Average Calibration Error (ACE) and the Area Under the Sparsification Error (AUSE) curve to report the calibration of the Bayesian models. Although the Expected Calibration Error \cite{guo2017calibration, naeini2015obtaining} is the \textit{de-facto} metric used in calibration of segmentation models, we follow \cite{neumann2018relaxed} to report the ACE, which is adapted for object detection models. The ACE score computes the absolute difference between the bin's accuracy and bin's confidence, giving equal weight to each bin and avoiding getting biased by the low-confidence predictions typical of object detectors. Finally, as standard calibration metric, we also report the Area Under the Sparsification Error curve (AUSE) \cite{ilg2018uncertainty} in terms of the Brier score \cite{brier} which provides a relative measure of the uncertainty.

\section{Results}

First, we show a comparative of our model with previous works, showing that Bayesian models outperform its deterministic version when they are correctly calibrated. We further perform an comparative study on the distribution of the dropout layers and its influence on the $F_\beta^w$, PMQ, ACE and AUSE scores and we show the evolution of the metrics with the number of samples $M$. Finally, we illustrate some qualitative results and the variance maps produced by the Bayesian model.

\begin{table*}[]
\centering
\resizebox{\textwidth}{!}{
\begin{tabular}{|l|c||c|c|c|cccccc|}
\hline
 & \begin{tabular}[c]{@{}c@{}}$F_w^b$ ($\%$) ($\uparrow$)\\ Deterministic ($M = 1$) \end{tabular} 
 & \begin{tabular}[c]{@{}c@{}}$F_w^b$ ($\%$) ($\uparrow$)\\ Bayesian ($M = 24$) \end{tabular} 
 & AUSE ($\downarrow$) 
 & ACE ($\downarrow$) 
 & $PMQ$ ($\%$)($\uparrow$) 
 & $pPMQ$ ($\%$)($\uparrow$) 
 & $Q_s$ ($\%$)($\uparrow$) 
 & $Q_l$ ($\%$)($\uparrow$) 
 & $FG$ ($\%$)($\uparrow$) 
 & $BG$ ($\%$)($\uparrow$) \\ \hline
R50 Baseline & 84.4 & - & - & - & 10.8 & 13.7 & 4.2 & 94.2 & 28.7 & 16.5 \\ \hline
R50 MC Enc & 85.6 & 84.9 & 0.150 & 0.00410 & 35.7 & 46.6 & 28.9 & 94.3 & 58.5 & 52.3 \\
R50 MC Enc-FC & 85.7 & 85.9 & 0.132 & 0.00357 & 34.7 & 57.2 & 40.5 & 95.2 & 56.5 & 71.5 \\
R50 MC Enc-M & 85.3 & 84.1 & 0.133 & 0.00385 & 31.3 & 49.8 & 33.4 & 92.7 & 58.9 & 58.2 \\
R50 MC Enc-FC-M & 84.5 & 85.7 & 0.134 & 0.00356 & 35.5 & 55.1 & 40.6 & 93.4 & \textbf{64.6} & 63.3 \\
R50 MC FC-M & 83.1 & 81.8 & 0.188 & 0.00453 & 30.8 & 45.4 & 29.6 & 93.6 & 56.5 & 53.3 \\ \hline
Rx101 Baseline & 86.1 & - & - & - & 12.5 & 15.6 & 4.9 & \textbf{97.6} & 32.9 & 17.8 \\ \hline
Rx101 MC Enc & \textbf{86.8} & 86.3 & 0.134 & 0.00359 & 36.7 & 44.8 & 27.6 & 95.0 & 62.5 & 44.4 \\
Rx101 MC Enc-FC & 86.6 & \textbf{86.9} & 0.129 & \textbf{0.00309} & \textbf{37.7} & \textbf{66.8} & \textbf{52.8} & 95.5 & 61.2 & \textbf{78.4} \\
Rx101 MC Enc-M & 85.6 & 84.4 & 0.128 & 0.00377 & 32.6 & 53.9 & 38.8 & 93.1 & 60.2 & 60.5 \\
Rx101 MC Enc-FC-M & 86.4 & 86.6 & \textbf{0.122} & 0.00328 & 36.7 & 55.5 & 41.3 & 93.0 & \textbf{64.6} & 59.6 \\
Rx101 MC FC-M & 85.5 & 84.3 & 0.154 & 0.00410 & 31.1 & 48.4 & 34.4 & 94.2 & 61.4 & 54.1 \\ \hline
\end{tabular}
}
\caption{Comparative of the uncertainty error (AUSE, ACE) and the affordance perception metrics ($F_\beta^w$, PMQ). $Q_s =$ Spatial Quality, $Q_l =$ Label Quality, FG = Foreground Quality ($exp(-L_{FG})$), BG = Background Quality ($exp(-L_{BG})$)}
\label{tab:my-table}
\end{table*}

\subsection{Affordance perception}

We compare our results with the $F_\beta^w$ score on the IIT-Aff dataset with previous works in Table \ref{tab:baselines}. We obtain the best performance with a 86.9 $\%$ $F_\beta^w$ for the Resnext-101 MC Enc-FC configuration, which is a +2.5 p.p improvement with respect to the previous state-of-the-art. We achieve the best performance on each affordance classes, except the \textit{cut} and \textit{pound} categories. We attribute this improvement due to the higher modelling capacity of the Resnext-101 encoder and the better refinement of the mask contour produced by the bayesian accumulation of $M$ predictions.

We evaluate the probabilistic affordance segmentation using the novel PMQ in Table \ref{tab:my-table}, which reveals the semantic and spatial performance of the detectors. The differences on the PMQ metric between the probabilistic and deterministic version are greater than the reported by the $F_\beta^w$ score. For example, the deterministic R50 scores a 84.4 $\% F_\beta^w$ and the R50 Enc-FC a 85.9 $F_\beta^w$ (Table \ref{tab:baselines}), while the PMQ is respectively 10.8 and 34.7 (Table \ref{tab:my-table}). The novel PMQ highlights the advantage of Bayesian detections models over the deterministic baselines: the average of the $M$ masks generates a better approximation to the ground truth and scores a higher $Q_s$ quality. In addition, the $Q_s$ and $Q_l$ qualities indicate that our affordance segmentation models are more likely to fail \textit{where} it is located rather than \textit{what} can we do with that object. It shows the relevance of the spatial uncertainty contribution at affordance segmentation and indicates a potential improving margin in this area.


\begin{figure}
\centering
\includegraphics[width=0.49\columnwidth]{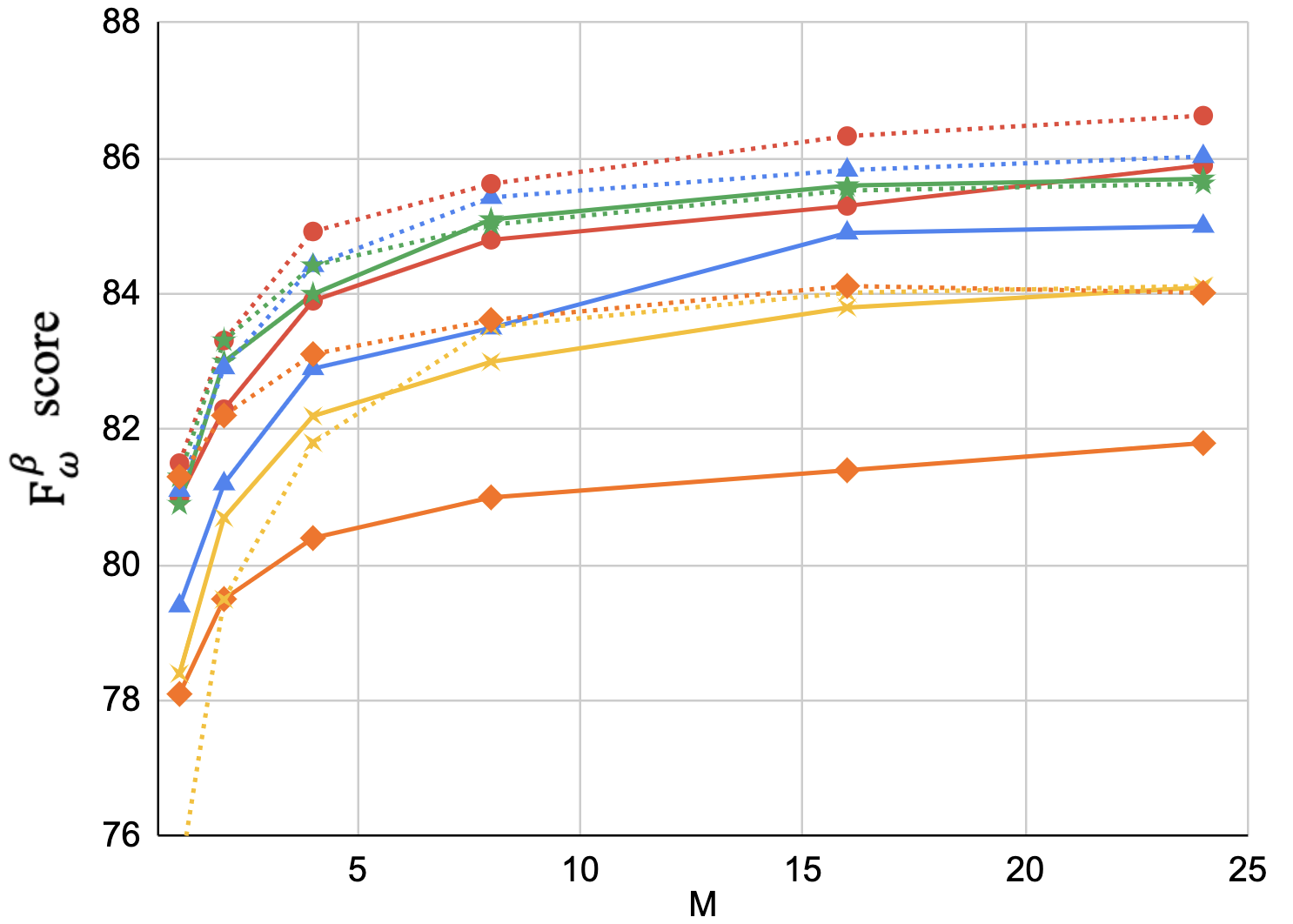} 
\includegraphics[width=0.49\columnwidth]{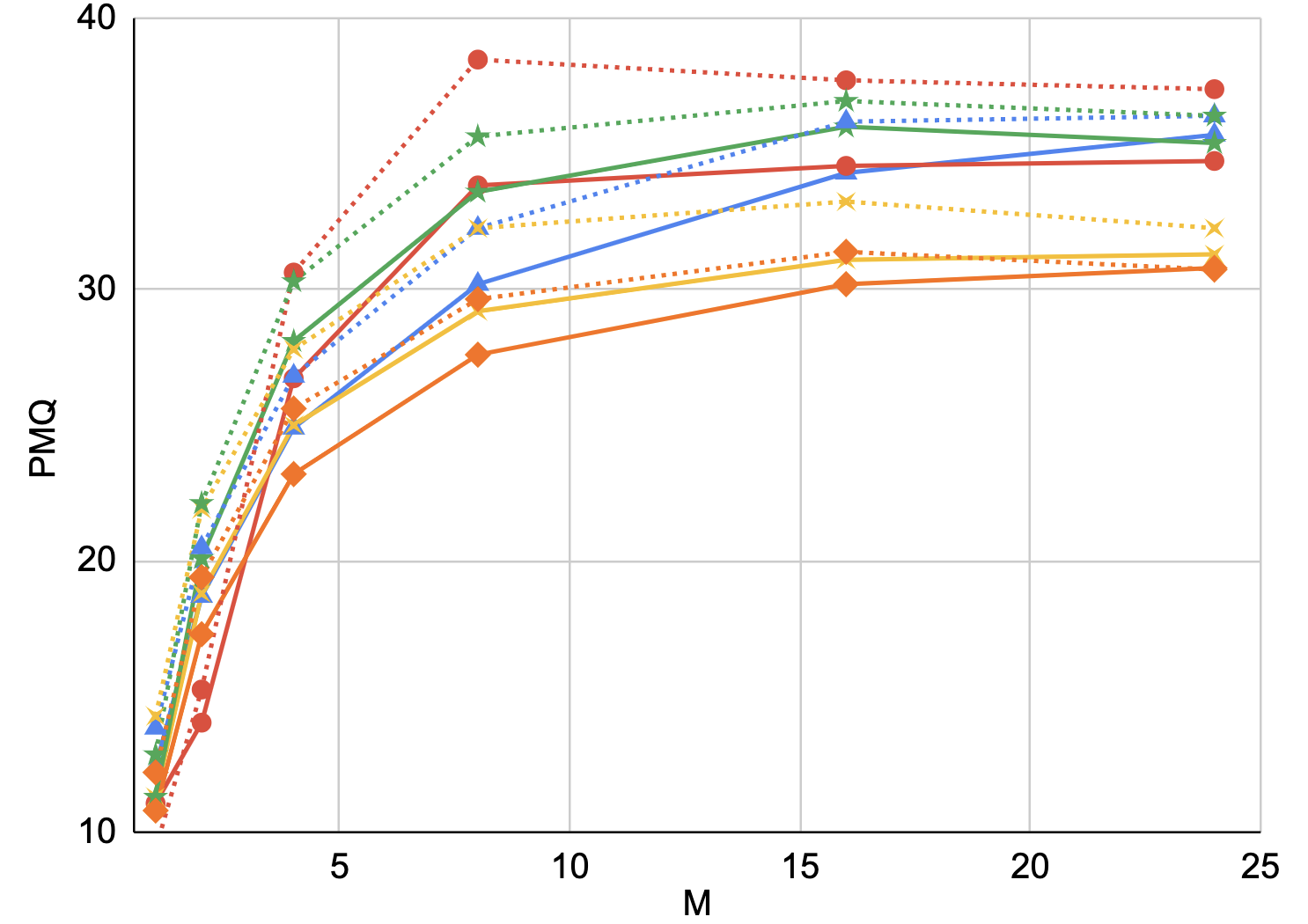} \\
\includegraphics[width=0.49\columnwidth]{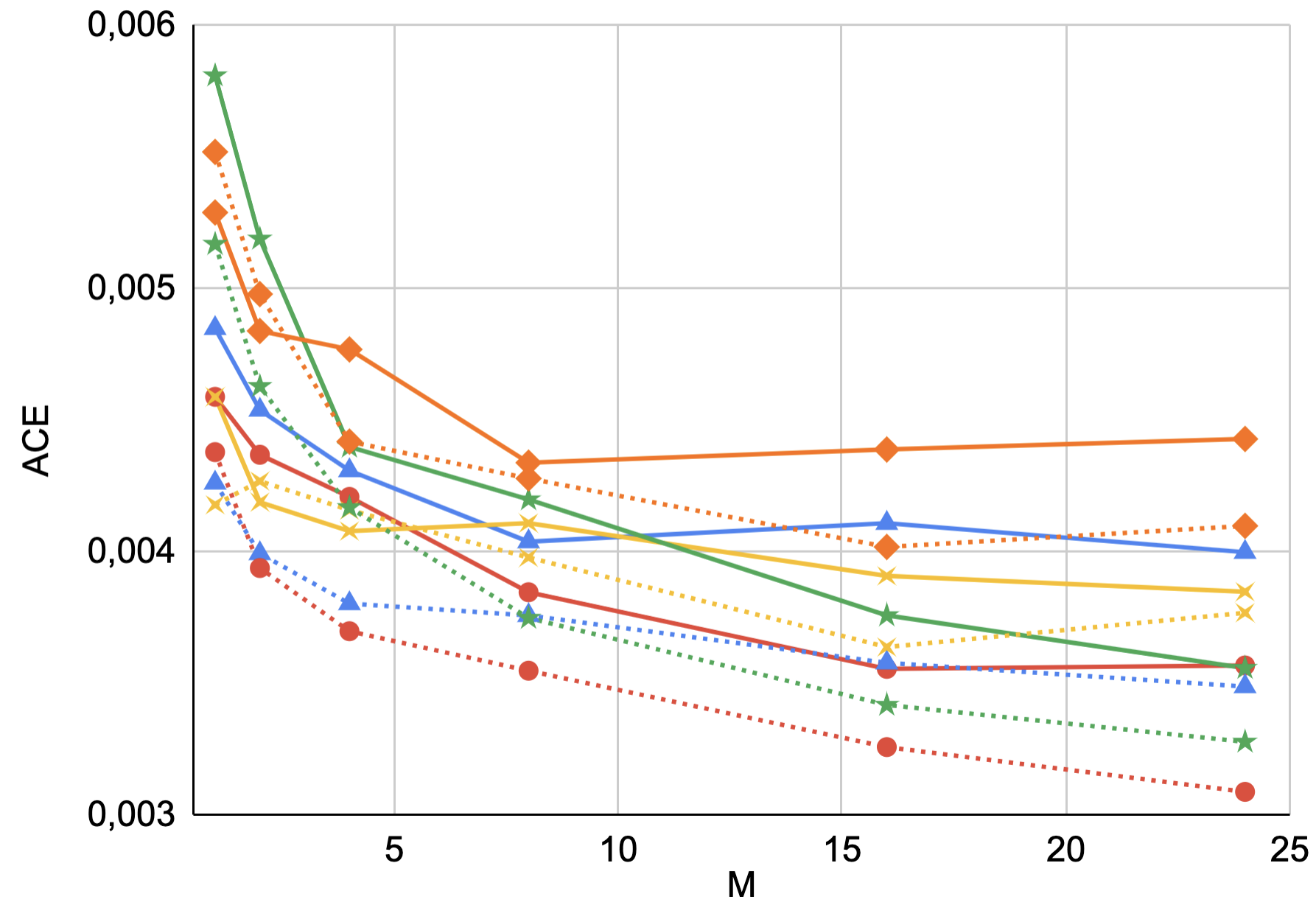} 
\includegraphics[width=0.49\columnwidth]{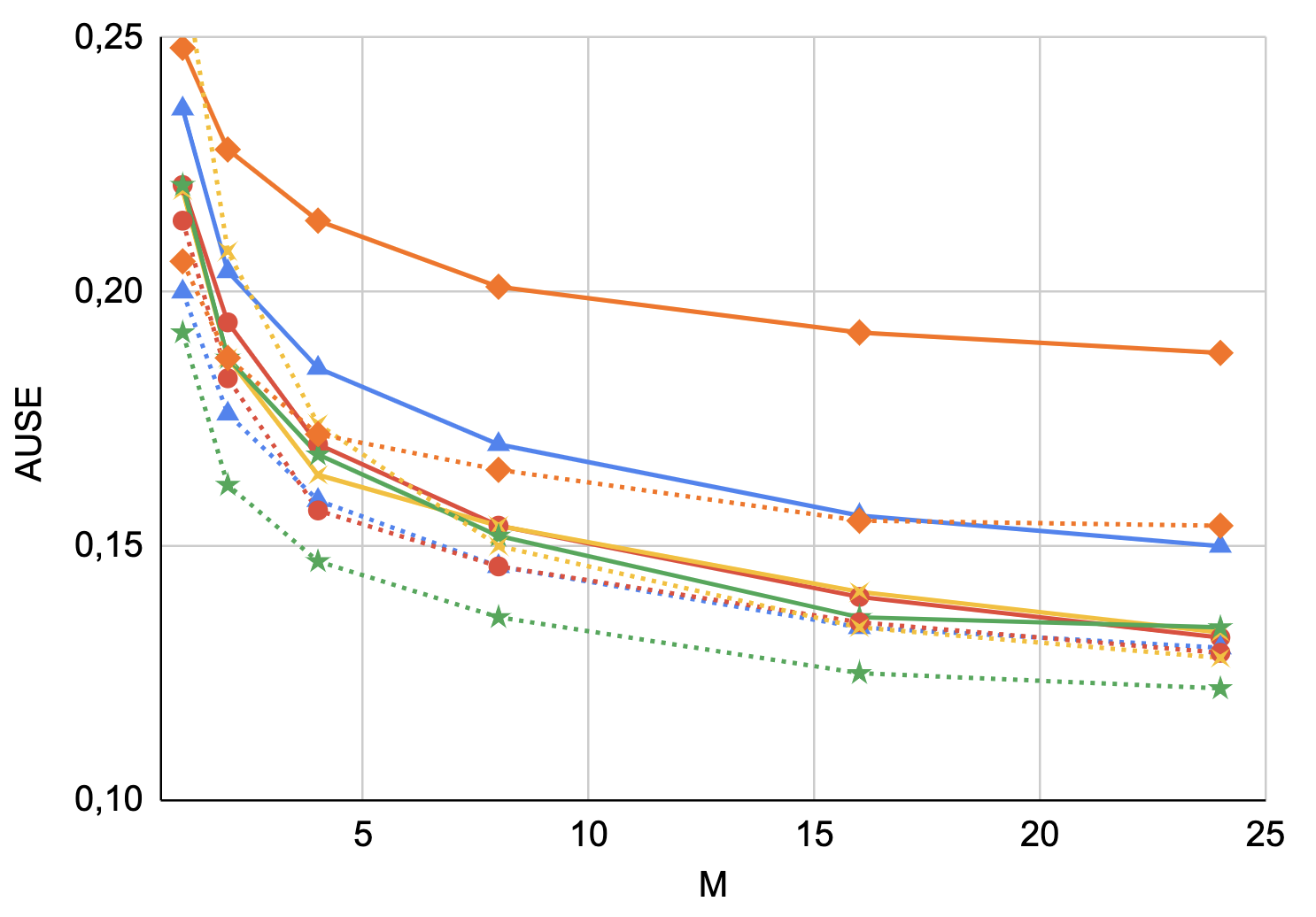} \\
\includegraphics[width=0.99\columnwidth]{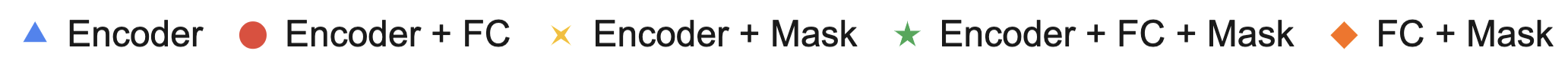}
\caption{Evolution with forward passes $M$ for different configurations of the affordance perception and uncertainty metrics. Top-left: $F_w^\beta$. Top-right: PMQ. Bottom-left: ACE. Bottom-right: AUSE. The metrics converge when the number of samples $M$ increase. The dotted lines represent the Rx101 version, while the R50 is represented by the continuous line.}
\label{fig:configurations_1}
\end{figure}

\begin{figure*}
\centering
\includegraphics[width=1.98\columnwidth]{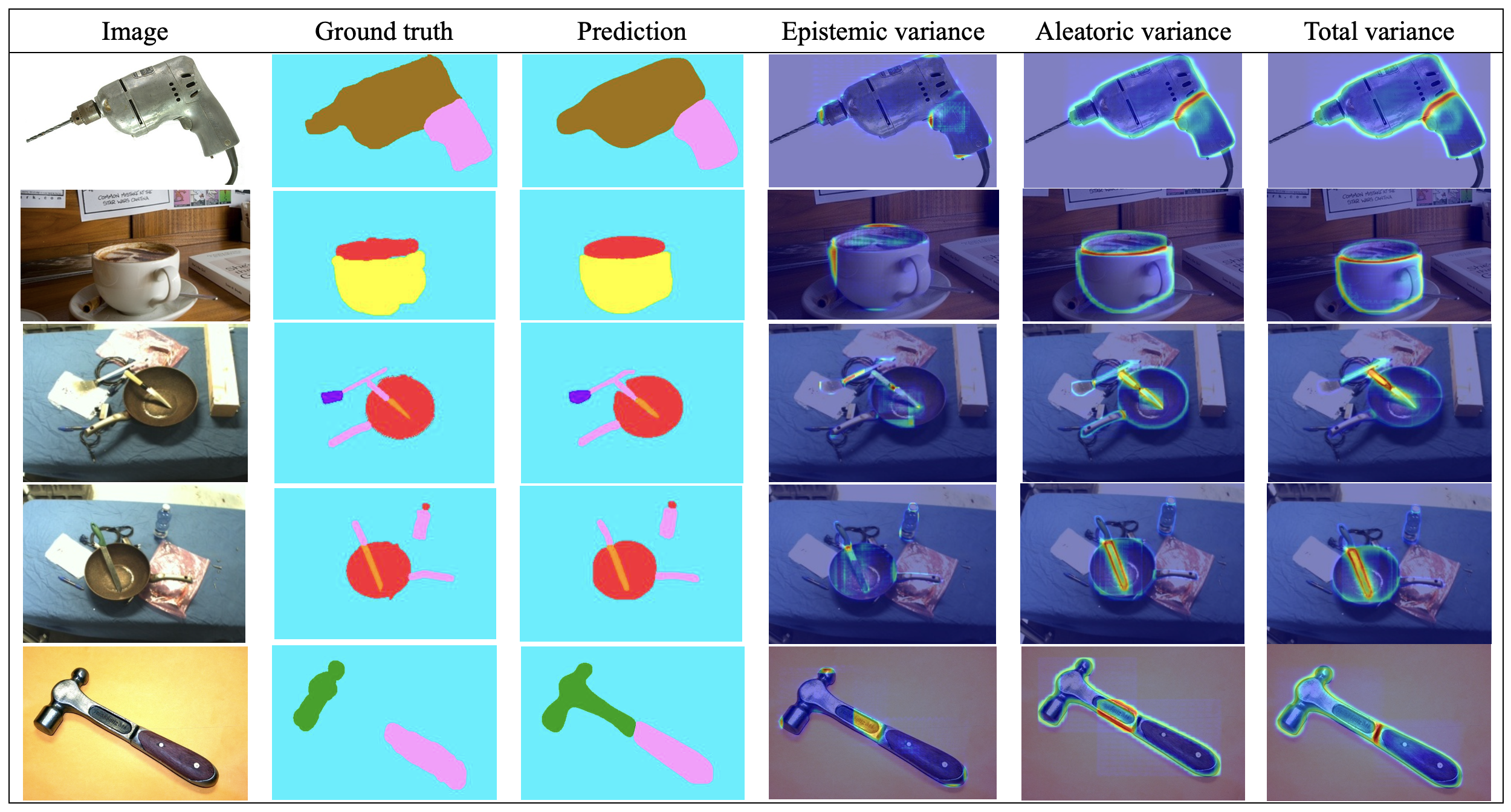} \\
\includegraphics[width=1.99\columnwidth]{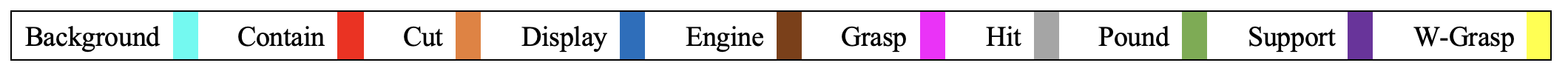} \\
\caption{Qualitative results. The predicted affordance masks by the Bayesian Rx101 are very smooth and match very closely with the ground truth. Epistemic variance appears in challenging pixels due to the presence of occlusions, light artefacts or ambiguous regions; while the aleatoric variance is in the contour of the objects due to the higher presence of noise.}
\label{fig:qualitative_results}
\end{figure*}

\subsection{Model analysis}
First of all, we set the dropout rate as $p_d = 0.5$ for both encoders using a grid search algorithm. An excessive dropout rate like $p_d = 0.7$ introduces too much variability, while $p_d = 0.1$ makes the model very close to the deterministic behaviour. 

\paragraph{The importance of the feature extraction modules} The modelling capacity in the latent space of the encoders shows significant performance differences. The R50 baseline (43.9 M trainable parameters) obtains 84.4 $\%$ $F_\beta^w$, compared with the Rx101 baseline that scores a 86.1 $\%$ $F_\beta^w$ (107.1 M parameters), as Table \ref{tab:baselines} shows. The PMQ metric also reflects this difference, where the Rx101 configurations overpass their respective R50 version in all the cases. For instance, Rx101 MC Enc-FC achieves a 37.7 PMQ, 66.8 pPMQ, 52.8 $Q_s$ and 95.5 $Q_l$; while the same R50 version obtains a 34.7 PMQ, 57.2 pPMQ, 40.5 $Q_s$ and 95.2 $Q_l$. In the same way, as Table \ref{tab:my-table} and Figure \ref{fig:configurations_1} show, the Rx101 models score lower uncertainty error metrics, showing that they are better calibrated and that their estimated probability is closer to the actual value. 

\paragraph{Influence of the distribution of the dropout layers} The localization of the dropout layers determines the model performance due to the different dropout effects. When we place the dropout at the backbone, it regularizes the largest part of the net and enforces the backbone to learn probabilistic representations in the latent space, producing a better-calibrated model with lower error metrics (Table \ref{tab:my-table}). The models with dropout activated at the backbone (MC Enc, MC Enc+FC, MC Enc+M, MC Enc+FC+M) surpass the performance of the configuration without dropout active at this stage (MC FC-M). The FC-M version presents the worst performance in all the metrics (84.3 $\%$ $F_\beta^w$ , 0.154 AUSE, 0.00410 ACE, 31.1 $\%$ PMQ on the Rx101 backbone) since the model learns deterministic representations due to the largest part of the architecture (the backbone) is still deterministic; but in the end on the architecture, we enforce it to produce variability and the uncertainty is not properly modelled in the feature space.

On the other hand, when we place the dropout at the FC-Mask layers we introduce the variability needed for the Bayesian prediction that induces later the uncertainty estimation, showing the differences between the Bayesian and deterministic behaviour for the same configuration. The results show that dropout on the FC layers is enough to propagate the variability to the final mask prediction. The dropout at the FC branch induces variability at the class label and the localization of the bounding box $b$, which translates lat to the mask branch. For example, the R50 MC Enc-FC-M improves from 84.5 to 85.7 on the  $\%$ $F^\beta_w$. Furthermore, placing dropout layers only at the convolutional layers of the Mask branch is detrimental for both the $F^\beta_w$ and PMQ scores, illustrated in Table \ref{tab:my-table}. An adequate configuration of the dropout calibrates better the Bayesian models and translates this improvement to the PMQ and $F^\beta_w$ scores.

\paragraph{Evolution with the number of samples $M$} We show in Figure \ref{fig:configurations_1} the evolution of the $\% F^\beta_w$, PMQ, AUSE and ACE metrics with the number of averaged samples $M$. First, we observe that the predictions improve with the number of averaged models $M$, but the improvement rate slows for $M > 16$. In the Bayesian models, as we increase $M$, the average of the predictions covers the loss induced by a single sample and the random dropout of $p_d$ neurons. According to the rest of the results, the metrics evolution of the Rx101 versions (dotted lines) overpasses its respective R50 (continuous line) configuration.

\subsection{Qualitative results}

We show a selection of the affordance masks predicted and the distribution of the epistemic and aleatoric uncertainty in Figure \ref{fig:qualitative_results}. Examining the inferred masks, they match very closely with the ground truth and they do not show any pixel artefacts. Due to the averaging of the $M$ samples for each detection, the final masks are very smooth and cover the totality of the affordance region. The method works in any type of scene, no matter the number of objects or the distance to the observer. As the differences between the $Q_l$ and the $Q_s$ show, there is not any semantically wrong detection, while the model encounters more difficulty in the spatial distribution of the masks. The spatial variance maps also show relevant information to the observer and offer insights into the model reasoning. The aleatoric variance is present in the contours of the objects due to the higher presence of noise and the difficulty of the model to produce a sharp contour, obtaining the highest values at the intersection between two contours. Epistemic variance appears in samples out of the distribution or visually challenging pixels. For example, the hammer handle in the last row shows a failure example of affordance segmentation, although the epistemic uncertainty shows an increased value due to the difficulty of determining if that region can be used to pound or not.

\section{Conclusions}

We introduce a novel extension of an instance segmentation model, Mask-RCNN, to perform the task of a Bayesian probabilistic affordance segmentation. Our results show that different configuration of the dropout layers yields different calibration and performance, over-passing the performance of the deterministic version and achieving the state-of-the-art on the IIT-Aff dataset. The localization of the dropout at the backbone enforces the model to learn a probabilistic latent space which models better the uncertainty estimation, while the decoder stages generate the variability to refine better the mask contours. In addition, we obtain the spatial epistemic and aleatoric variance at the pixel level. We enhance previous works that reduce the uncertainty to a single value or do not distinguish between the aleatoric and epistemic variance contributions. We also introduce a PMQ metric that compares the probabilistic masks to compute the spatial quality. We hope that our work inspires future research in affordance reasoning.

{\small
\bibliographystyle{IEEEtran}
\bibliography{mybib}

\begin{thebibliography}{10}
\providecommand{\url}[1]{#1}
\csname url@samestyle\endcsname
\providecommand{\newblock}{\relax}
\providecommand{\bibinfo}[2]{#2}
\providecommand{\BIBentrySTDinterwordspacing}{\spaceskip=0pt\relax}
\providecommand{\BIBentryALTinterwordstretchfactor}{4}
\providecommand{\BIBentryALTinterwordspacing}{\spaceskip=\fontdimen2\font plus
\BIBentryALTinterwordstretchfactor\fontdimen3\font minus
  \fontdimen4\font\relax}
\providecommand{\BIBforeignlanguage}[2]{{%
\expandafter\ifx\csname l@#1\endcsname\relax
\typeout{** WARNING: IEEEtran.bst: No hyphenation pattern has been}%
\typeout{** loaded for the language `#1'. Using the pattern for}%
\typeout{** the default language instead.}%
\else
\language=\csname l@#1\endcsname
\fi
#2}}
\providecommand{\BIBdecl}{\relax}
\BIBdecl

\bibitem{gibson}
J.~J. Gibson, \emph{The ecological approach to visual perception: classic
  edition}.\hskip 1em plus 0.5em minus 0.4em\relax Psychology Press, 2014.

\bibitem{jamone2016affordances}
L.~Jamone, E.~Ugur, A.~Cangelosi, L.~Fadiga, A.~Bernardino, J.~Piater, and
  J.~Santos-Victor, ``Affordances in psychology, neuroscience, and robotics: A
  survey,'' \emph{IEEE Transactions on Cognitive and Developmental Systems},
  vol.~10, no.~1, pp. 4--25, 2016.

\bibitem{nguyen2016detecting}
A.~Nguyen, D.~Kanoulas, D.~G. Caldwell, and N.~G. Tsagarakis, ``Detecting
  object affordances with convolutional neural networks,'' in \emph{2016
  IEEE/RSJ International Conference on Intelligent Robots and Systems
  (IROS)}.\hskip 1em plus 0.5em minus 0.4em\relax IEEE, 2016, pp. 2765--2770.

\bibitem{nguyen2017object}
A.~Nguyen, D.~Kanoulas, D.~G. Caldwell, and N.~Tsagarakis, ``Object-based
  affordances detection with convolutional neural networks and dense
  conditional random fields,'' in \emph{2017 IEEE/RSJ International Conference
  on Intelligent Robots and Systems (IROS)}.\hskip 1em plus 0.5em minus
  0.4em\relax IEEE, 2017, pp. 5908--5915.

\bibitem{do2018affordancenet}
T.-T. Do, A.~Nguyen, and I.~Reid, ``Affordancenet: An end-to-end deep learning
  approach for object affordance detection,'' in \emph{2018 IEEE international
  conference on robotics and automation (ICRA)}.\hskip 1em plus 0.5em minus
  0.4em\relax IEEE, 2018, pp. 5882--5889.

\bibitem{minh2020learning}
C.~N.~D. Minh, S.~Z. Gilani, S.~M.~S. Islam, and D.~Suter, ``Learning
  affordance segmentation: An investigative study,'' in \emph{2020 Digital
  Image Computing: Techniques and Applications (DICTA)}.\hskip 1em plus 0.5em
  minus 0.4em\relax IEEE, 2020, pp. 1--8.

\bibitem{caselles2021standard}
H.~Caselles-Dupr{\'e}, M.~Garcia-Ortiz, and D.~Filliat, ``Are standard object
  segmentation models sufficient for learning affordance segmentation?''
  \emph{arXiv preprint arXiv:2107.02095}, 2021.

\bibitem{Gal}
A.~Kendall and Y.~Gal, ``What uncertainties do we need in {Bayesian} deep
  learning for computer vision?'' \emph{arXiv preprint arXiv:1703.04977}, 2017.

\bibitem{gustafsson2020evaluating}
F.~K. Gustafsson, M.~Danelljan, and T.~B. Schon, ``Evaluating scalable bayesian
  deep learning methods for robust computer vision,'' in \emph{Proceedings of
  the IEEE/CVF conference on computer vision and pattern recognition
  workshops}, 2020, pp. 318--319.

\bibitem{feng2021labels}
D.~Feng, Z.~Wang, Y.~Zhou, L.~Rosenbaum, F.~Timm, K.~Dietmayer, M.~Tomizuka,
  and W.~Zhan, ``Labels are not perfect: Inferring spatial uncertainty in
  object detection,'' \emph{IEEE Transactions on Intelligent Transportation
  Systems}, 2021.

\bibitem{miller2021uncertainty}
D.~Miller, N.~S{\"u}nderhauf, M.~Milford, and F.~Dayoub, ``Uncertainty for
  identifying open-set errors in visual object detection,'' \emph{IEEE Robotics
  and Automation Letters}, vol.~7, no.~1, pp. 215--222, 2021.

\bibitem{harakeh2020bayesod}
A.~Harakeh, M.~Smart, and S.~L. Waslander, ``Bayesod: A bayesian approach for
  uncertainty estimation in deep object detectors,'' in \emph{2020 IEEE
  International Conference on Robotics and Automation (ICRA)}.\hskip 1em plus
  0.5em minus 0.4em\relax IEEE, 2020, pp. 87--93.

\bibitem{morrison2019uncertainty}
D.~Morrison, A.~Milan, and E.~Antonakos, ``Uncertainty-aware instance
  segmentation using dropout sampling,'' in \emph{Proceedings of the Robotic
  Vision Probabilistic Object Detection Challenge (CVPR 2019 Workshop), Long
  Beach, CA, USA}, 2019, pp. 16--20.

\bibitem{he2017mask}
K.~He, G.~Gkioxari, P.~Doll{\'a}r, and R.~Girshick, ``Mask r-cnn,'' in
  \emph{Proceedings of the IEEE international conference on computer vision},
  2017, pp. 2961--2969.

\bibitem{gal2016dropout}
Y.~Gal and Z.~Ghahramani, ``Dropout as a bayesian approximation: Representing
  model uncertainty in deep learning,'' in \emph{international conference on
  machine learning}.\hskip 1em plus 0.5em minus 0.4em\relax PMLR, 2016, pp.
  1050--1059.

\bibitem{rodriguez2022bayesian}
J.~Rodr{\'\i}guez-Puigvert, R.~Mart{\'\i}nez-Cant{\'\i}n, and J.~Civera,
  ``Bayesian deep neural networks for supervised learning of single-view
  depth,'' \emph{IEEE Robotics and Automation Letters}, vol.~7, no.~2, pp.
  2565--2572, 2022.

\bibitem{miller2019evaluating}
D.~Miller, F.~Dayoub, M.~Milford, and N.~S{\"u}nderhauf, ``Evaluating merging
  strategies for sampling-based uncertainty techniques in object detection,''
  in \emph{2019 International Conference on Robotics and Automation
  (ICRA)}.\hskip 1em plus 0.5em minus 0.4em\relax IEEE, 2019, pp. 2348--2354.

\bibitem{miller2018dropout}
D.~Miller, L.~Nicholson, F.~Dayoub, and N.~S{\"u}nderhauf, ``Dropout sampling
  for robust object detection in open-set conditions,'' in \emph{2018 IEEE
  International Conference on Robotics and Automation (ICRA)}.\hskip 1em plus
  0.5em minus 0.4em\relax IEEE, 2018, pp. 3243--3249.

\bibitem{kwon2020uncertainty}
Y.~Kwon, J.-H. Won, B.~J. Kim, and M.~C. Paik, ``Uncertainty quantification
  using bayesian neural networks in classification: Application to biomedical
  image segmentation,'' \emph{Computational Statistics \& Data Analysis}, vol.
  142, p. 106816, 2020.

\bibitem{lin2014microsoft}
T.-Y. Lin, M.~Maire, S.~Belongie, J.~Hays, P.~Perona, D.~Ramanan,
  P.~Doll{\'a}r, and C.~L. Zitnick, ``Microsoft coco: Common objects in
  context,'' in \emph{European conference on computer vision}.\hskip 1em plus
  0.5em minus 0.4em\relax Springer, 2014, pp. 740--755.

\bibitem{thermos2017deep}
S.~Thermos, G.~T. Papadopoulos, P.~Daras, and G.~Potamianos, ``Deep
  affordance-grounded sensorimotor object recognition,'' in \emph{Proceedings
  of the IEEE Conference on Computer Vision and Pattern Recognition}, 2017, pp.
  6167--6175.

\bibitem{myers2015affordance}
A.~Myers, C.~L. Teo, C.~Ferm{\"u}ller, and Y.~Aloimonos, ``Affordance detection
  of tool parts from geometric features,'' in \emph{2015 IEEE International
  Conference on Robotics and Automation (ICRA)}.\hskip 1em plus 0.5em minus
  0.4em\relax IEEE, 2015, pp. 1374--1381.

\bibitem{margolin2014evaluate}
R.~Margolin, L.~Zelnik-Manor, and A.~Tal, ``How to evaluate foreground maps?''
  in \emph{Proceedings of the IEEE conference on computer vision and pattern
  recognition}, 2014, pp. 248--255.

\bibitem{hall2020probabilistic}
D.~Hall, F.~Dayoub, J.~Skinner, H.~Zhang, D.~Miller, P.~Corke, G.~Carneiro,
  A.~Angelova, and N.~S{\"u}nderhauf, ``Probabilistic object detection:
  Definition and evaluation,'' in \emph{Proceedings of the IEEE/CVF Winter
  Conference on Applications of Computer Vision}, 2020, pp. 1031--1040.

\bibitem{guo2017calibration}
C.~Guo, G.~Pleiss, Y.~Sun, and K.~Q. Weinberger, ``On calibration of modern
  neural networks,'' in \emph{International conference on machine
  learning}.\hskip 1em plus 0.5em minus 0.4em\relax PMLR, 2017, pp. 1321--1330.

\bibitem{naeini2015obtaining}
M.~P. Naeini, G.~Cooper, and M.~Hauskrecht, ``Obtaining well calibrated
  probabilities using bayesian binning,'' in \emph{Twenty-Ninth AAAI Conference
  on Artificial Intelligence}, 2015.

\bibitem{neumann2018relaxed}
L.~Neumann, A.~Zisserman, and A.~Vedaldi, ``Relaxed softmax: Efficient
  confidence auto-calibration for safe pedestrian detection,'' 2018.

\bibitem{ilg2018uncertainty}
E.~Ilg, O.~Cicek, S.~Galesso, A.~Klein, O.~Makansi, F.~Hutter, and T.~Brox,
  ``Uncertainty estimates and multi-hypotheses networks for optical flow,'' in
  \emph{Proceedings of the European Conference on Computer Vision (ECCV)},
  2018, pp. 652--667.

\bibitem{brier}
G.~W. Brier \emph{et~al.}, ``Verification of forecasts expressed in terms of
  probability,'' \emph{Monthly weather review}, vol.~78, no.~1, pp. 1--3, 1950.

\end{thebibliography}
}

\end{document}